\documentclass[twoside,leqno,twocolumn]{article}

\usepackage[letterpaper]{geometry}

\usepackage{ltexpprt}
\usepackage{hyperref}

\usepackage[square,numbers]{natbib}
\usepackage{amsmath}
\usepackage{soul}
\usepackage{xcolor}
\usepackage[export]{adjustbox}
\usepackage{listings}
\usepackage{bm}
\usepackage{array,multirow,graphicx}
\usepackage{diagbox}
\usepackage[
  separate-uncertainty = true,
  multi-part-units = repeat
]{siunitx}
\usepackage{enumitem}
\usepackage{makecell}
\usepackage{tabularx}

\usepackage{mathtools}
\usepackage[T1]{fontenc}
\usepackage{algorithm}
\usepackage{algorithmic}

\usepackage{lipsum}
\let\OLDthebibliography\thebibliography
\renewcommand\thebibliography[1]{
  \OLDthebibliography{#1}
  \setlength{\parskip}{0pt}
  \setlength{\itemsep}{0pt plus 0.3ex}
}

\begin{document}

\newcommand\relatedversion{}

\title{\Large Message Propagation Through Time: An Algorithm for Sequence Dependency Retention in Time Series Modeling\relatedversion}

\author{Shaoming Xu \thanks{University of Minnesota.  \{xu000114, khand035,renga016,kumar001\}@umn.edu}
\and Ankush Khandelwal \footnotemark[1]
\and Arvind Renganathan \footnotemark[1]
\and Vipin Kumar \footnotemark[1]
}

\date{}
\maketitle




\fancyfoot[R]{\scriptsize{Copyright \textcopyright\ 2023\\
Copyright for this paper is retained by authors}}



\begin{abstract}
Time series modeling, a crucial area in science, often encounters challenges when training Machine Learning (ML) models like Recurrent Neural Networks (RNNs) using the conventional mini-batch training strategy that assumes independent and identically distributed (IID) samples and initializes RNNs with zero hidden states. The IID assumption ignores temporal dependencies among samples, resulting in poor performance. This paper proposes the Message Propagation Through Time (MPTT) algorithm to effectively incorporate long temporal dependencies while preserving faster training times relative to the stateful solutions. MPTT utilizes two memory modules to asynchronously manage initial hidden states for RNNs, fostering seamless information exchange between samples and allowing diverse mini-batches throughout epochs. MPTT further implements three policies to filter outdated and preserve essential information in the hidden states to generate informative initial hidden states for RNNs, facilitating robust training. Experimental results demonstrate that MPTT outperforms seven strategies on four climate datasets with varying levels of temporal dependencies.
\end{abstract}

\section{Introduction}
Machine learning (ML) models have achieved remarkable success in commercial applications such as computer vision and natural language processing, prompting the scientific community to consider them as viable alternatives to physics-based models \cite{willard2022integrating, legaard2023constructing}. Time series modeling, which seeks to uncover patterns, trends, and complex relationships in time-based data, is critical to scientific discovery \cite{lim2021time}. To train ML models on long time series, a common practice is to divide the time series into shorter sequence samples to reduce computational complexity and memory usage, and mitigate vanishing/exploding gradient issues \cite{pascanu2013difficulty, wen2022transformers}. 

However, the conventional random mini-batch algorithms treat these equal-sized sequences as IID and initialize RNNs with zero hidden states, leading to a loss of dependencies between the sequences. As a result, the trained RNNs cannot capture long-term temporal dependencies across multiple sequences and thus have limited performance for modeling long time series \cite{bengio1993problem, xu2023mini}. 

State-based approaches have been proposed to use hidden states to transfer temporal information across sequences. Stateful RNNs \cite{gulli2017deep} pass hidden states between mini-batches, but they treat sequences within each mini-batch as independent, necessitating smaller mini-batch sizes to maintain long-term dependencies \cite{yilmaz2019effect,elsworth2020time,katrompas2021enhancing}. This results in unstable training and extended training time \cite{bottou2007tradeoffs}. The Sequential Stateful RNNs \cite{xu2023mini} can preserve long-term dependencies by passing hidden states both within and between mini-batches. This allows larger mini-batches for stable model training. However, it requires each sequence within the batch to be processed sequentially, which makes it still slow to train. 

Response-based approaches use response values as additional inputs to encode temporal dependencies between sequences \cite{xu2023mini, williams1989learning, bengio2015scheduled}. These approaches are derived from ideas in dynamical control theory to adaptively let RNNs estimate responses as the initial system conditions for each sequence \cite{williams1989learning, abarbanel2013predicting, abarbanel2018machine, voss2004nonlinear, mikhaeil2022difficulty}. However, response-based approaches may suffer error accumulation during inference, as they primarily focus on learning responses' local changes in each timestep, impeding the learning of accumulated changes over multiple steps \cite{xu2023mini}. Moreover, they may struggle when responses encode limited temporal information.

To address these challenges, we propose a novel algorithm, Message Propagation Through Time (MPTT), which preserves long-term temporal dependencies by seamlessly transferring informative hidden states between sequences during training. 
First, MPTT utilizes two memory modules to enable asynchronous communication of hidden states between sequences. This approach supports shuffled sequences and diverse mini-batches across epochs, preserving long-term dependencies while improving computational efficiency relative to state-based approaches. 
Second, MPTT incorporates three policies to generate informative initial hidden states by filtering outdated information and retaining essential information in the hidden states throughout epochs. This facilitates more robust training than the stateful RNN approaches. 
Finally, MPTT addresses the error accumulation issue found in response-based approaches by utilizing informative hidden states, which possess a greater capacity than response values to encode the temporal effects and learn cumulative changes across timesteps. We highlight the advantage of these components in modeling long time series across four climate datasets with diverse temporal dependency levels. We have made our codes, datasets, and pre-trained models publicly available \footnote{\url{https://github.com/XuShaoming/Message-Propagation-Through-Time}} for further exploration and use.

\begin{figure*}[tb]
\centering
\includegraphics[width=\linewidth]{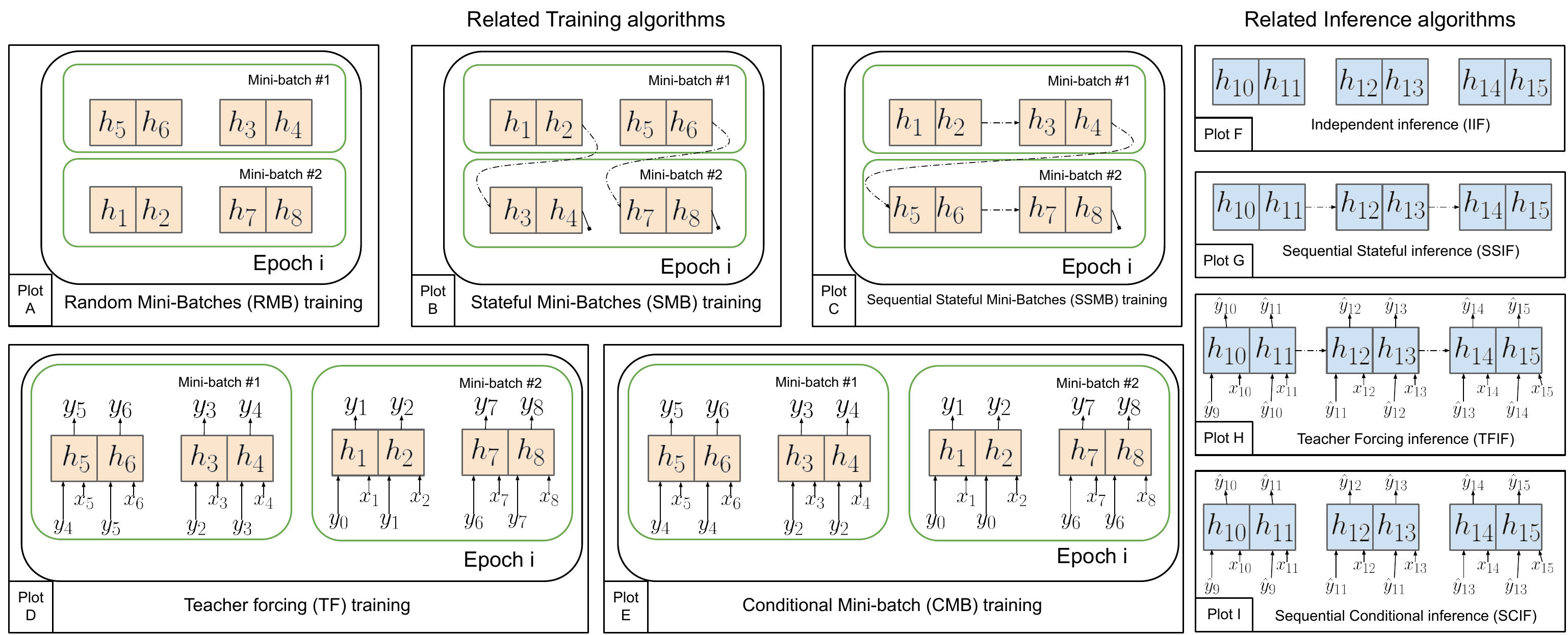} 
\caption{The related training and inference algorithms feed forward information flow between sequences, while backpropagation of losses across sequences is disallowed. For clarity, the sequence length is set to two, and inputs and outputs are omitted in some plots.}
\label{fig:MPTT_related}
\end{figure*}
\begin{figure*}[tb]
\centering
\includegraphics[width=\linewidth]{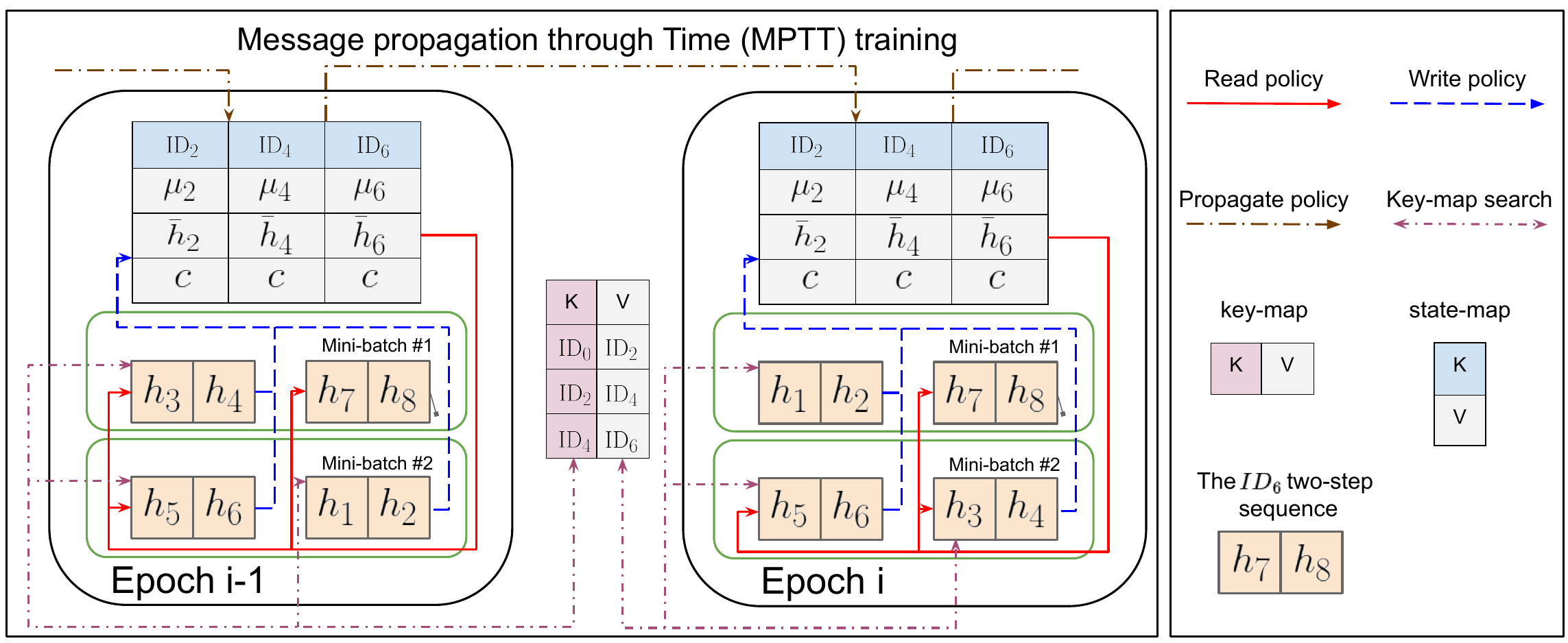}
\caption{The proposed Message Propagation through Time (MPTT) algorithm that designs two memory modules and three policies to generate informative initial hidden states to facilitate seamless communication between RNN sequences during training.}
\label{fig:MPTT}
\end{figure*}
\section{Background and Preliminaries}\label{sec:related_work}
\subsection{Problem Statement.} The goal is to accurately predict a sequence of outputs from the given time series of inputs. We primarily focus on the many-to-many prediction; however, the many-to-one is also possible but entails higher computational overhead. The overarching aim is to develop effective algorithms to transfer information across sequences.
\vspace{-0.5\baselineskip}
\subsection{Independent training and inference.} \hfill

\textbf{Random Mini-Batches (RMB) training} (Figure \ref{fig:MPTT_related} plot A) is a common mini-batch gradient descent method \cite{ruder2017overview} for RNNs, which involves dividing longer time series into shorter sequences and treating each sequence independently. In each epoch, RMB randomly allocates sequences to multiple mini-batches, initializes the hidden states to $\vec{0}$ for each sequence, and trains the models on each mini-batch until all mini-batches have been processed. 

\textbf{Independent Inference (IIF)} (Figure \ref{fig:MPTT_related}, plot F) is the inference algorithm that considers each test sequence IID and makes predictions using $\vec{0}$ as initial states. The combination of RMB and IIF algorithms represents a widely-used learning strategy in the ML community. However, it neglects the interactions between sequences, potentially leading to suboptimal performance in time series modeling. Both RMB and IIF generate RNNs' hidden states following the equation:
\begin{align} \label{eq:IIF_seq}
\begin{split}
h_t &= f(x_t, h_{t-1}) = f(x_t, f(x_{t-1}, f(x_{t-2}, \dots f(x_{1}, \vec{0})\dots)))
\end{split}
\end{align}
where $x$ denotes the input sequence, and $x_t$ and $h_t$ denote the data and hidden state at the time step t of the sequence, respectively. $W$ denotes the length of the sequence, and $0<t<W$. Next, we describe existing strategies to incorporate temporal dependencies in RNNs.
\subsection{State-based approaches}
\vspace{-0.5\baselineskip}
aim to transfer temporal information between sample sequences through the initial hidden state $h_0$ as follows:
\begin{align} \label{eq:State-based}
\begin{split}
h_t &= f(x_t, h_{t-1}) = f(x_t, f(x_{t-1}, f(x_{t-2}, \dots f(x_{1}, h_0)\dots)))
\end{split}
\end{align}
In equation \ref{eq:State-based}, $h_0$ represents the hidden state one step before the current input sequence $x$ and is computed as the last state of the previous sequence using the same model $f$. In the implementation, $h_0$ is detached from the computational graph to prevent error backpropagation over sequences and reduce computational cost. The fundamental idea behind state-based approaches is that initial hidden states can summarize the temporal effects of previous sequences, thereby preserving sequence dependencies.

\textbf{Stateful Mini-Batches (SMB) training} (Figure \ref{fig:MPTT_related} plot B) uses the last hidden states of RNN sequences from a mini-batch to initialize the sequences in the next mini-batch to train Stateful RNNs \cite{gulli2017deep}. However, SMB still treats each sequence within the mini-batch as independent, limiting their ability to capture long-term dependencies. Consequently, researchers often have to compromise by using smaller mini-batch sizes to link more sequences for applications with long-term dependencies \cite{yilmaz2019effect,elsworth2020time,katrompas2021enhancing}, which can result in a prolonged and unstable training process \cite{bottou2007tradeoffs}.

\textbf{Sequential Stateful Mini-Batches (SSMB) training} (Figure \ref{fig:MPTT_related} plot C) can preserve long-term dependencies while allowing model training on larger mini-batches for stable training by passing hidden states both within and between mini-batches. However, the fixed sequence order required by both SMB and SSMB for hidden state transmission can potentially lead to overfitting. Moreover, SSMB requires additional computational cost, as it processes each sequence individually to pass hidden states sequentially \cite{xu2023mini}.

\textbf{Sequential Stateful inference (SSIF)} (Figure \ref{fig:MPTT_related}, plot G) serves as the default inference algorithm for the state-based training algorithms. It passes hidden states between test sequences according to their temporal order to generate predictions in a single pass. As SSIF prevents RNNs from reconstructing hidden states from scratch for each sequence during inference, SSIF can also improve the predictions of RMB-trained models \cite{xu2023mini}.

\subsection{Response-based approaches}
\vspace{-0.5\baselineskip}
encode the temporal relationship between sequences through response variables, akin to ideas in dynamical control theory to adaptively estimate responses as the system conditions for predictions \cite{williams1989learning, abarbanel2013predicting, abarbanel2018machine, voss2004nonlinear, mikhaeil2022difficulty}. These approaches employ previous response values as additional inputs, aiding the model in learning temporal variations.
  
\textbf{Teacher Forcing training (TF)} (Figure \ref{fig:MPTT_related}, plot D) 
utilizes the observed response value from the previous timestep as an additional input for the next timestep, as shown in equation \ref{eq:TF} \cite{williams1989learning}. TF uses the initial response value $y_0$ from the previous sequences to maintain dependencies between the current and previous sequences. However, TF struggles with time-series modeling because it focuses on learning local changes relative to previous response values, leading to error accumulation in the inference phase \cite{xu2023mini}.

\textbf{Scheduled sampling (SSPL) training} \cite{bengio2015scheduled} mitigates the exposure bias of TF by incrementally replacing the observed responses with the predictions as additional inputs during training. However, SSPL does not solve TF's inability to learn accumulated changes in time series modeling.

\textbf{Teacher Forcing inference (TFIF)} (Figure \ref{fig:MPTT_related}, plot H) serves as the inference algorithm for TF and SSPL trained RNNs. TFIF incorporates the previous timestep's predicted response as additional input for the next timestep \cite{williams1989learning}. However, since TF and SSPL cannot learn accumulated changes during training, any errors made by TFIF during inference can accumulate over time, leading to poor time-series predictions.
\begin{align} \label{eq:TF}
\begin{split}
    h_t &= f(x_t, y_{t-1}, h_{t-1}) = f(x_t,y_{t-1},\\
    &f(x_{t-1},y_{t-2}, f(x_{t-2},y_{t-3}, \dots f(x_1, y_0, \vec{0})\dots)))
\end{split}
\end{align}

\textbf{Conditional Mini-Batches (CMB) training} (Figure \ref{fig:MPTT_related}, plot E) mitigates the error accumulation issue in TF and SSPL by using a single response value as an initial condition for each sequence, specifically, the response value one timestep before the start of each training sequence. This initial response value is replicated across the sequence, maintaining the input vector's dimensionality and offering the corresponding RNN sequence a starting point to learn accumulated changes over time (as shown in equation \ref{eq:CMB_hs}) \cite{xu2023mini}. CMB is inspired by the multiple shooting method in control theory, where the entire time series is divided into sequences, and an initial condition is estimated for each sequence\cite{voss2004nonlinear, mikhaeil2022difficulty}. CMB can effectively train RNNs for applications characterized by long-term dependencies, as their responses encapsulate the temporal effects of previous timesteps. However, for applications where the response variables encapsulate limited temporal information, CMB is not better than RMB \cite{xu2023mini}.

\textbf{Sequential conditional inference (SCIF)} (Figure \ref{fig:MPTT_related}, plot I) serves as the inference algorithm for CMB. SCIF maintains the temporal order of sequences and sequentially uses the initial response predicted from the previous sequence as the additional input for  CMB-trained models at the current sequence to obtain all predictions in one pass \cite{xu2023mini}.
\begin{align} \label{eq:CMB_hs}
\begin{split}
    h_t &= f(x_t, y_0, h_{t-1}) 
    = f(x_t,y_0,\\
    &f(x_{t-1},y_0, f(x_{t-2},y_0,\dots f(x_1, y_0, \vec{0})\dots)))
\end{split}
\end{align}
\section{Proposed Approach} \label{sec:MPTT}
The primary objective of the MPTT algorithm is to maintain the computational efficiency of mini-batch training while enforcing temporal dependencies between sequences. The core idea in MPTT is to use informative hidden states, denoted as $\tilde{h}_0$, as the messages to transmit temporal information among sequences as follows:
\begin{align} \label{eq:MPTT_hs}
\begin{split}
    h_t &= f(x_t, h_{t-1}) = f(x_t, f(x_{t-1}, f(x_{t-2}, \dots f(x_{1}, \tilde{h}_0)\dots)))
\end{split}
\end{align}
Next, we outline the memory modules and policies that MPTT employs to manage message propagation throughout the training procedure.

\subsection{Memory modules}
\vspace{-0.5\baselineskip}
In state-based approaches,  subsequent sequences must await the generation of hidden states from preceding ones to initiate training, which dictates the sequential nature of the approaches and can lead to overfitting and suboptimal training \cite{xu2023mini,gulli2017deep}. MPTT overcomes these issues by employing two memory modules, \emph{key-map} and \emph{state-map}, to enable asynchronous reading and writing of messages as the informative initial hidden states for RNNs, as illustrated in Figure \ref{fig:MPTT}. This facilitates seamless information sharing between sequences, allowing for shuffled sequences and diverse mini-batches to address overfitting and local minima during mini-batch training.

\textbf{Key-map}.
MPTT assigns each sequence $S$ a unique ID:\{$T$\}, where $T$ denotes the initial step of $S$. The initial step refers to the timestep one step before $S$ in the original time series. For instance, a sequence with ID:\{365\} starts at timestep 366 in the time series.

The key-map comprises key-value pairs, where each key corresponds to the ID of a specific sequence $S$, and its value is a list of IDs representing the sequences whose initial steps are covered in $S$. A smaller stride of the sliding window can result in more overlapped sequences, leading to longer lists in the key-map. The key-map can be represented as follows:
\begin{align} \label{eq:key_map}
\begin{split}
\textbf{key-map}(\text{ID}_i) = [\text{ID}_j | \forall T_j, T_i < T_j \leq T_i + W]
\end{split}
\end{align}
where $W$ is the sequence length, and $\text{ID}_i$ and $T_i$ represent $S_i$ and its initial step, respectively.

Before starting the training process, a key-map is initialized for each sequence using Equation \ref{eq:key_map} and remains fixed without new training data. The key-map serves as a mediator that enables a given sequence $S$ to identify the following sequences that require its generated hidden states for initializing their hidden states.

\textbf{State-map}. \label{sec:state_map}
We can represent the state-map as
\begin{align} \label{eq:state_map}
\begin{split}
\textbf{state-map}(\text{ID}) = (\mu_0, \bar{h}_0, c),
\end{split}
\end{align}
The state-map maintains three variables, $(\mu_0, \bar{h}_0, c)$, for each sequence $S$, which are used to generate the initial message $\tilde{h}_0$ for initializing the hidden states. $\mu_0$ encapsulates information from a collection of initial hidden states, $h_0$'s, produced in prior epochs for a sequence $S$. $\bar{h}_0$ represents a running average of $h_0$'s generated during the current epoch, while $c$ denotes the count of $h_0$'s in the current epoch for $S$. Before the start of training, a state-map is initialized for each sequence using Equation \ref{eq:state_map}, but with $(\Vec{0}, \vec{0}, 0)$ as the assigned value. The state-map functions as a repository for each sequence, storing pertinent information needed to generate the messages $\tilde{h}_0$'s as initial hidden states.
\subsection{Policies}\label{sec:mptt_policies}
\vspace{-0.5\baselineskip}
MPTT employs read, write, and propagate policies to maintain the message $\tilde{h}_0$ and its $\{\mu_0, \bar{h}_0, c\}$ for each sequence $S$ during training. 
In each epoch, MPTT randomly partitions the sequences into multiple mini-batches without replacement. It then iterates through each mini-batch to train the model.

\textbf{Read policy}. For each sequence in a mini-batch, MPTT reads its $\{\mu_0, \bar{h}_0, c\}$ from the \emph{state-map} and generates $\tilde{h}_0$ as follows:
\begin{align} \label{eq:MPTT_fetch_policy}
    \tilde{h}_0 = 
    \begin{cases}
    \mu_0 & \text{if } \delta=0, c=0 \\
    \frac{\delta \mu_0  + c\bar{h}_0}{\delta + c} & \text{Otherwise}
    \end{cases}
\end{align}
The binary value  $\delta = \{0,1\}$ is called the message keeper. If $\delta = 0$, \emph{read policy} will disregard  $\mu_0$ as soon as $\bar{h}_0$ is updated in the current epoch. If $\bar{h}_0$ has not been updated ($c=0$), \emph{read policy} will always utilize $\mu_0$ as the initial hidden state for RNNs to generate the prediction $\hat{y}$ and hidden state $h$ at each timestep of the sequence $S$. The \emph{read policy} sets initial hidden states for each sequence at the beginning of the mini-batch. 

\textbf{Write Policy}. After the optimization step, each sequence $S$ in the current mini-batch uses the \emph{key-map} to identify corresponding sequences that require its generated hidden states for their \emph{state-map} updates. For a sequence $\text{ID}_j$ in the key-map of the sequence $\text{ID}_i$, the hidden state generated during the forward pass at the $T_j - T_i$ timestep of sequence $\text{ID}_i$ is used as $h_0$ to update the \emph{state-map} for sequence $\text{ID}_j$ as follows:
\begin{align} \label{eq:MPTT_update_policy}
\begin{split}
    \mu_0=\mu_0, \quad \bar{h}_0 &= \frac{c\bar{h}_0 + h_0}{c+1}, \quad c = c + 1
\end{split}
\end{align}

\textbf{Propagate policy}. The \emph{Propagate Policy} is executed at the end of each epoch, generating a new $\mu_0$ and resetting both $\bar{h}_0$ and $c$ in the \emph{state-map} for each sequence as follows:
\begin{align} \label{eq:MPTT_pass_policy}
    \mu_0 &= \frac{\delta\mu_0 + c \bar{h}_0}{\delta+c}, \qquad  \bar{h}_0 = \vec{0}, \qquad  c=0
\end{align}
If message keeper $\delta=0$, \emph{propagate policy} will disregard $\mu_0$ derived from previous epochs and rely solely on $\bar{h}_0$ from the current epoch as the new $\mu_0$ to initialize hidden states in the succeeding epoch. 

\subsection{Forgetting mechanism}
\vspace{-0.5\baselineskip}
We can expand equation \ref{eq:MPTT_pass_policy} and check the $\mu_0$ in each epoch $E$ as follows:
\begin{align} \label{eq:forgetting_mechanism}
\begin{split}
\mu_0 = 
\begin{cases}
\vec{0} & \text{if } E=1 \\
\bar{h}_0^{<1>} &\text{if } E=2 \\
\frac{\delta^{E-2}\bar{h}_0^{<1>}}{(\delta+c)^{E-2}} + \sum_{e=3}^{E} \frac{\delta^{E-e}c\bar{h}_0^{<e-1>}}{(\delta+c)^{E-e+1}} & \text{if } E > 2
\end{cases}
\end{split}
\end{align}
Where $E>2$ case is summarized from this expanded form:
\begin{align} \label{eq:forgetting_mechanism_expanded} 
\begin{split} 
    \mu_0 &= \frac{\delta^{E-2}\bar{h}_0^{<1>}}{(\delta+c)^{E-2}} + \frac{\delta^{E-3}c\bar{h}^{<2>}}{(\delta+c)^{E-2}} + \frac{\delta^{E-4}c\bar{h}^{<3>}}{(\delta+c)^{E-3}} + \\
    &\dots + \frac{\delta c\bar{h}^{<E-2>}}{(\delta+c)^{2}} + \frac{c\bar{h}^{<E-1>}}{(\delta+c)}
\end{split}
\end{align}
Here, $\bar{h}_0^{<e>}$ is the average of $h_0$'s, and  $c$ is the total count of the $h_0$'s generated at the end of the epoch $e$ for a sequence $S$. These equations reveal that MPTT utilizes three types of forgetting mechanisms to control the messages propagating over epochs.

\textbf{Message keeper $\delta$}. 
MPTT employs the message keeper $\delta=\{0,1\}$ to control the amount of information propagated. When $\delta=0$, the \emph{read policy} disregards $\mu_0$ as soon as $\bar{h}_0$ has been updated in the current epoch. The \emph{propagate policy} discards $\mu_0$ derived from previous epochs and solely relies on $\bar{h}_0^{<E-1>}$ as the new $\mu_0$ for the epoch $E$.

\textbf{Exponential decay}.
Conversely, when $\delta=1$, both the \emph{read} and \emph{propagate} policies utilize the full list of $\bar{h}_0$'s generated from previous epochs to obtain $\tilde{h}_0$ and $\mu_0$. Equation \ref{eq:forgetting_mechanism_expanded} shows that this process is regulated by exponential decay on the low-quality $\bar{h}_0$'s generated by under-trained RNNs in early epochs, prioritizing $\bar{h}_0$'s from more recent epochs to obtain new $\mu_0$ for the subsequent epoch.

\textbf{Auto-adjusted forgetting}.
Equation \ref{eq:forgetting_mechanism_expanded} underscores how the decay speed is automatically adjusted by the value of $c$. At the end of each epoch, $c$ corresponds to the total number of $h_0$'s generated for each sequence $S$. A larger $c$ indicates more overlapping sequences and an augmented dataset for training the RNNs, resulting in a wider array of $h_0$'s that are generated and aggregated to yield $\bar{h}_0$ with reduced variance in each epoch. Consequently, MPTT employs equation \ref{eq:forgetting_mechanism_expanded} to accelerate the decay process when $c$ is larger, thereby emphasizing the importance of recently generated $\bar{h}_0$'s in producing new $\mu_0$.
\section{Results and Discussion}\label{sec:Empirical_evaluation}
\begin{table*}[htbp]
\caption{The table displays experiment results where the sequence length is fixed at 366 steps. Models are trained using different percentages of the 500-year training set and evaluated on the testing set to measure their relative performance in RMSE. The results indicate that, when provided with more data, MPTTs reach lower RMSEs more quickly than other approaches, highlighting their ability to effectively leverage extra data for robust time-series modeling.}
\vspace{-1\baselineskip}
\begin{center}
\resizebox{\linewidth}{!}{
\begin{tabular}{c c |c c c c c|c c c c c|c c c c c}
\hline
\multicolumn{2}{c|}{\backslashbox{\textbf{Algo}}{\textbf{Data}}}&\multicolumn{5}{c|}{\textbf{SWAT-SW}}& \multicolumn{5}{c|}{\textbf{SWAT-SNO}}& \multicolumn{5}{c}{\textbf{SWAT-SF}}\\
\hline
\textbf{Train} & \textbf{Inference}& \textbf{2\%}& \textbf{8\%} & \textbf{16\%} & \textbf{32\%} & \textbf{100\%} & \textbf{2\%}& \textbf{8\%} & \textbf{16\%} & \textbf{32\%} & \textbf{100\%} & \textbf{2\%} & \textbf{8\%} & \textbf{16\%} & \textbf{32\%} & \textbf{100\%}\\
\hline
RMB& IIF&41.1&40.3&36.9&34.8&32.8&7.24&5.05&4.15&3.14&2.57&1.52&1.01&0.89&0.79&0.66\\
\hline
RMB& SSIF&36.4&34.1&27.3&23.7&20.1&7.13&4.76&3.75&2.39&1.67&1.49&0.94&0.81&0.7&0.53\\
\hline
TF& TFIF &35.4&40.1&27.9&26&27.7&\textbf{6.58}&\textbf{2.15}&\textbf{1.49}&\textbf{1.22}&1.61&1.63&0.96&0.83&0.81&0.67\\
\hline
SSPL& TFIF &49.3&87.1&34.8&30&34.1&8.14&2.59&1.76&1.31&1.06&1.48&0.94&0.83&0.76&0.61\\
\hline
CMB& SCIF &35.6&32.2&23.1&19.1&16.4&6.96&4.01&2.61&1.95&1.07&1.61&1.02&0.9&0.81&0.68\\
\hline
SMB& SSIF &36&32.9&27.2&26.4&18.2&7.05&5.35&3.73&2.18&1.4&1.55&1.03&0.86&0.77&0.54\\
\hline
SSMB& SSIF &36.9&\textbf{31}&\textbf{22.6}&18.2&14&6.74&4&3.25&2.27&1.39&1.55&0.96&0.77&0.7&0.49\\
\hline
MPTT($\delta=0$)& SSIF &\textbf{35}&32.4&23.2&17.2&11.5&7.19&4.36&2.69&1.91&\textbf{1.04}&1.51&0.89&0.76&0.65&0.48\\
\hline
MPTT($\delta=1$)& SSIF &35.4&31.5&22.8&\textbf{14.1}&\textbf{10.9}&6.89&4.02&2.62&1.8&1.12&\textbf{1.38}&\textbf{0.88}&\textbf{0.71}&\textbf{0.62}&\textbf{0.44}\\
\hline
\end{tabular}
}
\label{tab:SWAT}
\end{center}
\vspace{-1\baselineskip}
\end{table*}
\begin{figure*}[tb]
\centering
\includegraphics[width=\linewidth]{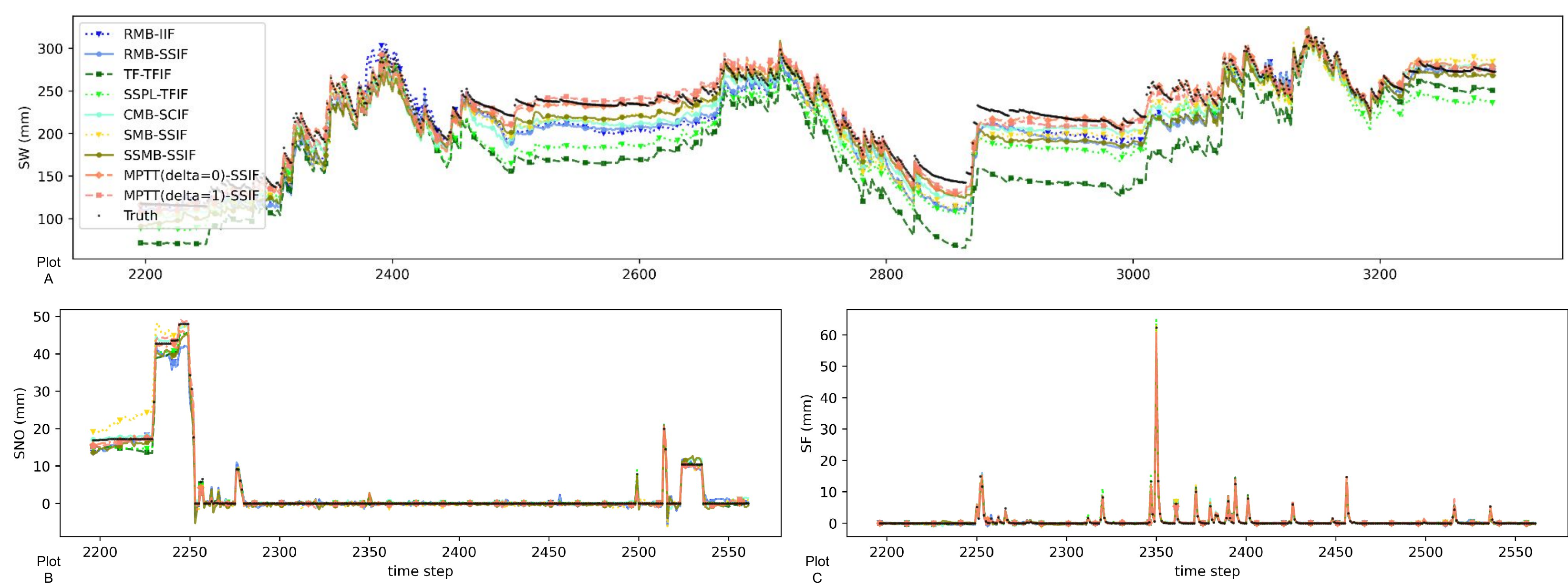} 
\caption{Comparisons between observed and predicted time series reveal algorithmic performance variations across datasets with different temporal dependencies. SW exhibits long-term dependencies, SNO has seasonal dependencies due to snowmelt, and SF features short-term dependencies driven by precipitation and snowmelt. The performance gap narrows from the SW to the SF tasks, indicating that MPTT algorithms excel particularly in scenarios with extended temporal dependencies.}
\label{fig:error_accumulate}
\end{figure*}
\begin{figure*}[htbp]
\centering
\includegraphics[width=\linewidth]{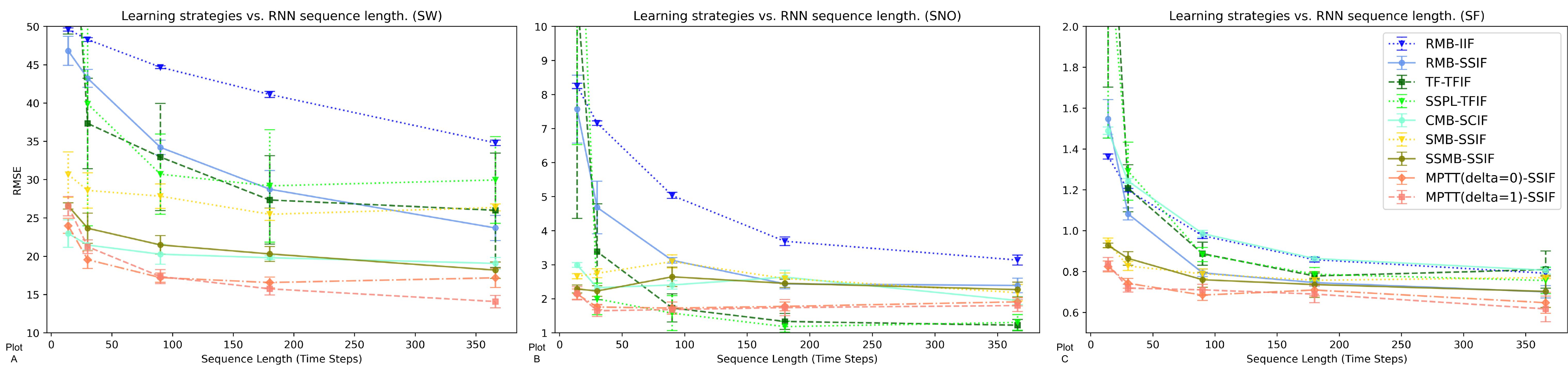} 
\caption{This figure displays experiment results where the training data size is fixed at 160 years. Models are trained with sequences having varying lengths (14, 30, 90, 180, 366 steps). Results indicate that all learning strategies perform better on longer sequences that retain more temporal dependencies.}
\label{fig:winsize}
\end{figure*}

We evaluate the performance of MPTT algorithms by comparing them with RMB, state-based, and response-based algorithms for training RNN models on four climate datasets. Combined with five inference algorithms used during testing, this leads to nine learning strategies for comparison.
\subsection{Synthetic data experiments} \label{sec:SWAT_exps}
\vspace{-0.5\baselineskip}
The Soil \& Water Assessment Tool (SWAT) is a river basin model for simulating surface and groundwater dynamics in complex watersheds and assessing climate change impacts \cite{arnold2012swat, bieger2017introduction}. In this experiment, we evaluate our algorithms by training the Gated Recurrent Units (GRU) models \cite{cho2014learning} as SWAT surrogate models. We generate three 1000-year daily-scale time series datasets for a Minnesota watershed using the SWAT model. These datasets, \emph{SWAT-SW}, \emph{SWAT-SNO}, and \emph{SWAT-SF}, represent soil water (SW), snowpack (SNO), and streamflow (SF) respectively. As Figure \ref{fig:error_accumulate} shows, SW has a long-term temporal dependency, reflecting soil water content changes. SNO exhibits a seasonal dependency, as snow typically melts by summer's end. SF, with a short-term temporal dependency, is mainly influenced by precipitation and snowmelt events. These responses depend on input weather drivers like precipitation, temperature, solar radiation, wind speed, and relative humidity (see appendix \ref{sec:appendix_SWAT_exps_data}).

\textbf{Experimental Setup.} We split the time series into 500 years of training, 100 years of validation, and 400 years of testing sets in temporal order. The data is Gaussian-normalized before being segmented into sequences, each with a length of 366 days. To evaluate strategy stability, we use RMB-IIF optimized hyperparameters (Table \ref{tab:hyperparameters}) to train five GRUs separately for each strategy. The performance is then evaluated based on the mean RMSE of the five GRUs for each strategy.
\begin{table}[ht]
    \caption{The RMB-IIF optimized hyperparameters.}
    \label{tab:hyperparameters}
    \centering
    \resizebox{\columnwidth}{!}{
    \begin{tabular}{|c|c|c|c|c|c|c|c|c|}
        \hline
         Model& loss &H&$\eta$& optimizer & B & E & dropout\\
         \hline
        GRU & mse&32 &0.01 & Adam  & 64& 500 & 0\\
        \hline
        LSTM & mse&256 &0.001 & Adam  & 64& 200& 0.4\\
        \hline
    \end{tabular}
    }
    {\small
        \resizebox{\columnwidth}{!}{
            \begin{tabular}{lcc}
                \textbf{mse}: mean square error & \textbf{H}: hidden state size & \textbf{B}: batch size\\
                \textbf{$\eta$}: learning rate & \textbf{E}: maximum epoch &
                
            \end{tabular}
        }
    }
\end{table}

\textbf{Impact of temporal dependency.} Table \ref{tab:SWAT} and Figure \ref{fig:error_accumulate} show that the algorithms perform differently on datasets with varying temporal dependencies.
Among response-based methods, TF and SSPL perform poorly on SW as they excessively rely on the previous timestep response, resulting in error accumulation and poor performance during inference. They perform nicely on SNO since complete snowmelt by summer's end naturally prevents error accumulation. CMB-SCIF performs well on SW and SNO, as their initial responses provide a solid foundation for accumulating changes over time. The three response-based approaches perform similarly to basic RMB-IIF on SF, as its short temporal dependency means responses carry limited information. 
MPTT algorithms perform nicely in modeling both SW and SF and reasonably well with SNO, largely due to their efficient utilization of hidden states for message generation.
This avoids error accumulation issues for variables with long-term dependencies and improves predictions for variables with short-term dependencies, where hidden states carry more information than responses.

\textbf{Impact of training sequence length.}
Figure \ref{fig:winsize} suggests that all learning strategies yield improved performance with longer sequences, which capture more temporal dependencies.
It further offers these insights:
(a) As temporal dependencies decrease (from SW to SF), the performance gap among algorithms narrows, underscoring our proposed strategies' effectiveness in utilizing temporal dependencies when present.
(b) Despite using the same RMB-trained models, RMB-SSIF consistently outperforms RMB-IIF, highlighting that even models trained using IID assumption can benefit from message passing between samples during inference. 
(c) The three response-based approaches exhibit distinct performances. CMB-SCIF performs well on SW and SNO. However, TF-TFIF and SSPL-TFIF perform poorly on SW, regardless of sequence length, and only perform well on SNO for longer sequences. All three strategies show no difference from RMB-IIF on SF predictions because of weak temporal auto-correlation in response values for streamflow.
(d) In contrast, state-based approaches, SMB-SSIF and SSMB-SSIF, perform well on varying sequence lengths, as they use hidden states during both training and inference phases, reducing the loss of temporal dependencies due to shorter sequences.
(e) MPTT algorithms consistently perform the best across all three datasets, regardless of sequence lengths, as they use high-quality hidden states and train models on heterogeneous mini-batches compared to state-based approaches.

\textbf{Time efficiency.}
Table \ref{tab:training_epoch_time} presents the per-epoch training time for various algorithms on the sequence length of 366 on an NVIDIA A100 GPU. Results show MPTT uses around 0.06 seconds/epoch, considerably faster than SSMB and preserving full sample and temporal dependencies. MPTT lags behind RMB by 0.04 seconds per epoch, a delay partly attributed to their Python dictionary-based memory modules, indicating room for optimization.
\begin{table}[hb]
    \caption{Per-epoch training time for various algorithms.}
    \label{tab:training_epoch_time}
    \resizebox{\linewidth}{!}{
    \begin{tabular}{c|c|c|c|c|c|c|c}
        \hline
         & RMB & TF & SSPL & CMB & SMB& SSMB  & MPTT\\
        \hline
        sec/epoch &0.0172&0.0217 &0.02 & 0.0151 & 0.0211& 0.539 & 0.0614\\
        \hline
    \end{tabular}
    }
\end{table}
\begin{table*}[tb] 
    \caption{The table displays the experiment results where each learning strategy is employed to train five separate LSTMs, utilizing hyperparameters outlined in Table \ref{tab:hyperparameters}. The final prediction for each basin in the CARAVAN test set is then obtained by averaging the predictions from these five LSTMs to evaluate the learning strategies.}
    \label{tab:CARAVAN}
    \resizebox{\linewidth}{!}{
    \begin{tabular}{c|c|c|c|c|c|c|c|c}
        \hline
         & RMB-IIF & RMB-SSIF & TF-TFIF & SSPL-TFIF & CMB-SCIF& SMB-SSIF  & MPTT($\delta=0$)-SSIF &MPTT($\delta=1$)-SSIF\\
        \hline
        RMSE &1.301&1.285 &1.465 & 1.273& 1.284& 1.368 & \textbf{1.255}&\textbf{1.255}\\
        \hline
        $R^2$ &0.674&0.686 &0.482 & 0.692 & 0.692& 0.625 & \textbf{0.705}&0.694 \\
        \hline
    \end{tabular}
    }
\end{table*}
\subsection{Real world data experiments.} \label{sec:caravan_uk}
\vspace{-0.5\baselineskip}
CARAVAN, a comprehensive global hydrology dataset, integrates data from seven large-scale hydrology studies \cite{kratzert2022caravan}. Long Short-Term Memory (LSTM) models has outperformed traditional models in rainfall-runoff predictions \cite{kratzert2019towards, gauch2021rainfall}. In this paper, we use nine dynamic meteorological forcings and 20 static basin characteristics to train LSTMs for predicting streamflow across 191 Great Britain (GB) basins in the CARAVAN dataset. Our results affirm the efficacy of our algorithms in modeling real-world complex systems.

\textbf{Experimental Settings.} 
We partition the time series into training (1989/10/01-1997/09/30), validation (1997/10/01-1999/09/30), and testing (1999/10/01-2009/09/30) periods. Data is normalized using the training set's mean and variance and then segmented into 365-day sequences with 183-day overlaps. Given the substantial variation in streamflow scales across basins, we use the Coefficient of Determination ($R^2$) in figures to evaluate predictions.

\begin{figure}[tb]
    \centering
    \includegraphics[width=\columnwidth]{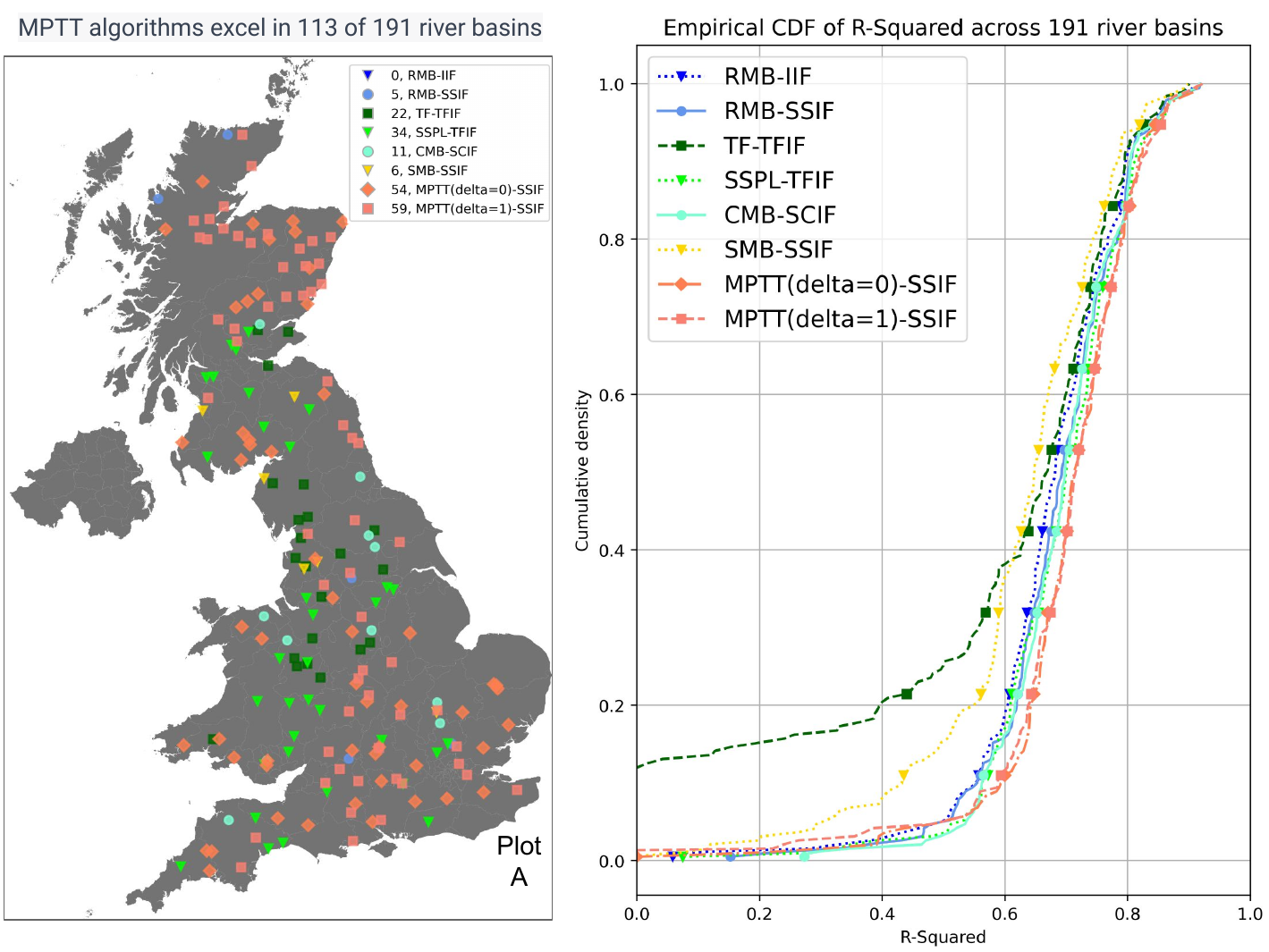}
    \caption{Plot A identifies the top-performing strategy for each basin. The results reveal that MPTT algorithms are the leading choice in 113 out of 191 basins. Plot B illustrates the empirical Cumulative Distribution Function (ECDF) of $R^2$ values. A curve leaning towards one on this graph suggests the model's effectiveness in explaining data variance across 191 basins. The results indicate that MPTT-based algorithms outperform their counterparts: only around 10\% of basins show an $R^2$ value below 0.6, compared to approximately 20\% for other algorithms.}
    \label{fig:carvan_cdf_map}
\end{figure}

\textbf{Top strategies for river basin streamflow prediction}. 
Table \ref{tab:CARAVAN} reveals that MPTT algorithms yield the best overall performance across all basins. Figure \ref{fig:carvan_cdf_map} Plot A specifically identifies the leading strategy for each river basin, leading to these observations: 
(a) RMB-IIF and RMB-SSIF only excel in two basins, while SMB-SSIF performs well in four. Their subpar performance underscores the importance of preserving sequence and long-term temporal dependencies and the need to avoid fixed mini-batches in training real-world applications.
(b) Response-based algorithms run relatively well, with TF-TFIF and SSPL-TFIF achieving the best predictions in 49 basins. 
(c) 
MPTT algorithms excel in 113 out of 191 basins. Within MPTT algorithms, MPTTs(delata=1) yield slightly more best-predicted basins than MPTTs(delata=0).

\textbf{A closer look at algorithm performance}. 
Figure \ref{fig:carvan_cdf_map}, Plot B, displays the ECDF curves for 191 basins and offers the following observations:
(a) Response-based approaches exhibit varied results; TF-TFIF has 22 best-predicted basins, but its 40\% basins have an $R^2$ lower than 0.6, while SSPL-TFIF and CMB-SCIF have only about 20\% with $R^2$ lower than 0.6. This suggests that TF-based algorithms have a large variance in their performance across 191 basins.
(b) SSPL-TFIF and CMB-SCIF display ECDF curves similar to RMB-IIF and RMB-SSIF, indicating no significant improvement in predictions. 
(c) MPTT-based algorithms outperform others, with almost 15\% of basins having an $R^2$ above 0.8, compared to approximately 10\% for other algorithms.

\section{Limitations and Future Directions} \label{sec:limitation_future}
The MPTT algorithm presents some limitations. 
First, MPTT is designed for state-based models; adapting it to transformers may require incorporating state information into the transformer architecture.
Second, while the fixed $\delta=1$ slightly outperforms $\delta=0$ in experiments, the optimal choice for MPTT message keeper $\delta$ is still uncertain, leaving room for future exploration.
Third, MPTT uses two memory modules, the key-map and state-map, which result in an additional asymptotic space complexity of $\Theta(M(H+\frac{W}{\Delta}))$, where  $M$ is the total number of sequences in the training set,$H$ is hidden state size, $W$ is the sliding window size, and $\Delta$ is the stride size. 
Fourth, MPTT is slower than RMB due to its frequent interactions with the memory modules. Despite potential optimizations, the extra computational cost is inherent to MPTT's design. 
Finally, this study highlights MPTT's strength in RNN training within environmental contexts. Future research could extend its applicability to diverse state-based models like Graph Neural Networks, aiming to improve optimization and information quality. Additionally, exploring MPTT's utility in sectors like healthcare, finance, and language processing offers promising avenues for expansion.

\section{Conclusion} \label{sec:conclusion}
This paper introduces novel MPTT algorithms designed to effectively capture sample and temporal dependencies when training sequential models for time series. MPTTs design two memory modules and three policies to generate informative initial hidden states as messages and facilitate seamless message communication between RNN sequences while allowing sample shuffling and diverse mini-batches across epochs during training. Experimental results show MPTTs consistently outperforming various competing algorithms on four climate datasets with multi-scale temporal dependencies, achieving the best predictions for 113 of 191 Great Britain river basins, demonstrating their effectiveness in modeling complex real-world dynamical systems. 

\section{Acknowledgement}
This work was funded by the National Science Foundation through the Harnessing the Data Revolution award (1934721) and the U.S. Department of Energy's Advanced Research Projects Agency–Energy (ARPA-E) award (DE-AR0001382). We sincerely appreciate Zhenong Jin, Licheng Liu, Xiang Li, Xiaowei Jia, Rahul Ghosh, Haoyu Yang, and Michael Steinbach for their insightful feedback. We are also thankful for the access to advanced computing resources provided by the Minnesota Supercomputing Institute.

\bibliographystyle{unsrt}
\bibliography{references.bib}
\appendix
\begin{center}
\section*{Appendix}
\end{center}

\section{Implementations} \label{appendix:Implementations}
\vspace{-0.5\baselineskip}
This section provides implementation details for some algorithms. For all algorithms, please refer to the code, data, and pre-trained models provided \footnote{\url{https://github.com/XuShaoming/Message-Propagation-Through-Time}}.

\subsection{Stateful Mini-Batches (SMB) training}
\vspace{-0.5\baselineskip}
We implement SMB to initialize the next mini-batch RNN sequences using the last hidden states of the current mini-batch. This necessitates adjacent, non-overlapping sequences. When dividing $M$ sequences by mini-batch size $B$, a remainder may arise. For example, in the synthetic data experiment with $M=498$ total sequences and a mini-batch size of $B=64$, we get seven mini-batches plus 50 remaining sequences. The first 50, along with 14 subsequent sequences, form the first mini-batch for RMB training. The residual 448 sequences are organized into seven mini-batches for SMB training.

\subsection{Sequential Stateful Mini-Batches (SSMB) training}
\vspace{-0.5\baselineskip}
In multi-entity datasets like CARAVAN-GB, which features 191 distinct river basins, SSMB and SMB function similarly. Since sequences from different entities are naturally independent, they can be conveniently grouped into a single mini-batch without requiring inter-entity information sharing.

\subsection{Scheduled Sampling (SSPL)}
\vspace{-0.5\baselineskip}
We employ the inverse sigmoid decay function to gradually reduce the probability of using observations as training inputs over epochs, represented by the decay rate $\epsilon_e$ in Equation \ref{eq:SSPL_decay}.
\begin{align} \label{eq:SSPL_decay}
\begin{split}
\epsilon_e = 
\begin{cases}
\frac{1}{1 + \exp{(\alpha \times (\frac{e}{E_{\text{decay}}} - \beta) )}} & \text{if } e\leq E_{\text{decay}} \\
0 & \text{if } E_{\text{decay}} < e \leq E \\
\end{cases}
\end{split}
\end{align}
In Equation \ref{eq:SSPL_decay}, $\alpha$ dictates the rate of decay, accelerating the reduction of ground truth usage with higher values. $\beta$ establishes the midpoint probability for using ground truth at 0.5. The variable $e$ signifies the current epoch, while $E_{\text{decay}}$ serves as the maximum decay epoch. The function is operational until $e \leq E_{\text{decay}}$, after which only predicted values are employed as inputs, setting $\epsilon_e = 0$.
\begin{table}[hb]
    \caption{Scheduled sampling settings}
    \label{tab:SSPL_decay}
    \resizebox{\columnwidth}{!}{
    \begin{tabular}{|c|c|c|c|c|}
        \hline
         Experiments & $E$ & $E_{\text{decay}}$ & $\alpha$ & $\beta$\\
        \hline
        Synthetic data &500& 400 &10 & 0.5\\
        \hline
        Real world data &200&150 &10 & 0.5\\
        \hline
    \end{tabular}
    }
\end{table}

\begin{figure}[htbp]
\centering
\includegraphics[width=\columnwidth]{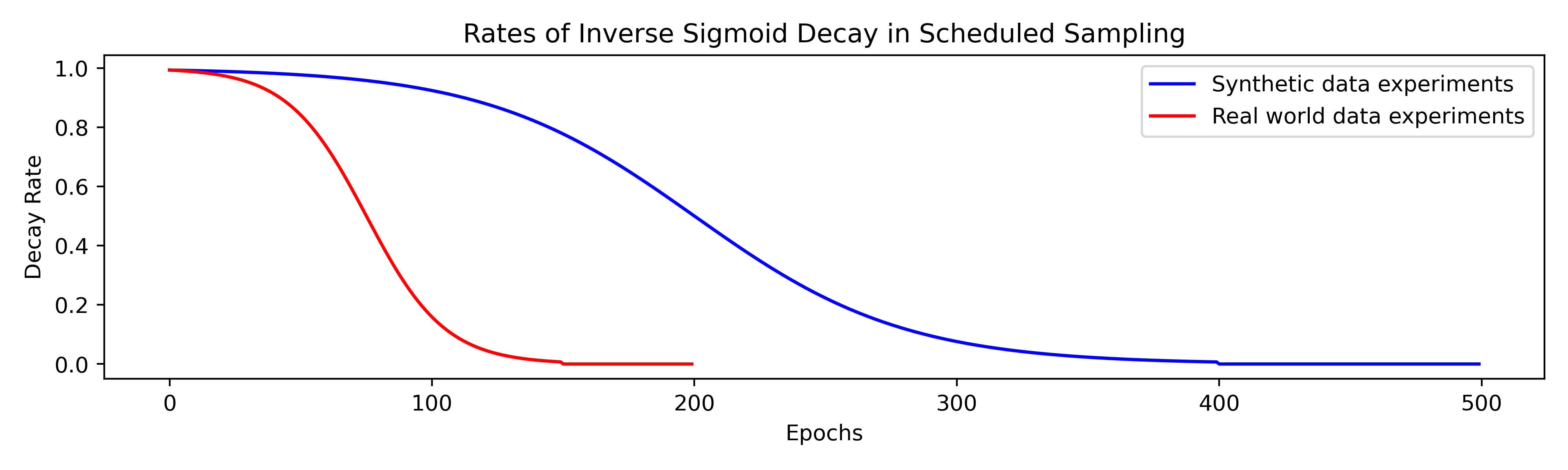} 
\caption{This figure depicts the inverse sigmoid decay rates for Scheduled Sampling (SSPL), as configured in Table \ref{tab:SSPL_decay}.}
\label{fig:SSPL_decay}
\end{figure}

Using the inverse sigmoid decay function, we transition the model to increasingly rely on its predictions, enhancing its adaptability during inference. Table \ref{tab:SSPL_decay} lists the hyperparameters controlling decay rates for synthetic and real-world experiments, while Figure \ref{fig:SSPL_decay} illustrates the epoch-wise decay trends. 

\subsection{Teacher Forcing inference (TFIF)}
\vspace{-0.5\baselineskip}
In TFIF inference, we permit sharing hidden states between adjacent sequences, thanks to RNNs that have uniform unit weights and can treat the hidden state from the adjacent sequence in the same way as the previous time step. This eliminates the need for RNNs to reconstruct hidden states and leads to stable AvgStepRMSE outcomes, as depicted in Figure \ref{fig:RMSEs_timestep}. 

\subsection{Sequential conditional inference (SCIF)}
\vspace{-0.5\baselineskip}
For SCIF inference, we refrain from transferring hidden states between sequences to maintain the integrity of the CMB-trained model. During training, CMB uses zeroed initial hidden states ($h_0$) and initial response values ($y_0$) as the starting point. Transferring hidden states between adjacent sequences in SCIF can compromise these starting points and confuse the CMB-trained model,  resulting in inaccurate predictions.

\section{Synthetic data experiments}\label{sec:appendix_SWAT_exps}
\subsection{Datasets} \label{sec:appendix_SWAT_exps_data}
\begin{figure}[htbp]
\centering
\includegraphics[width=\columnwidth]{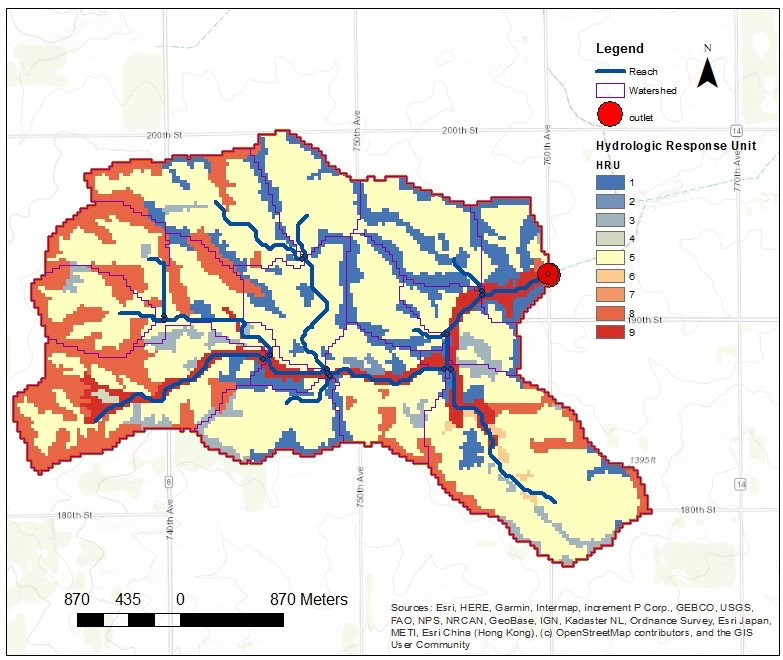} 
\caption{SWAT model setup for the South Branch of the Root River Watershed, MN, USA. }
\label{fig:SWAT_model}
\end{figure}
The \emph{SWAT-SW}, \emph{SWAT-SNO}, and \emph{SWAT-SF} datasets target soil water (SW), snowpack (SNO), and streamflow (SF) variables, derived from the Soil \& Water Assessment Tool (SWAT) \cite{arnold2012swat, bieger2017introduction} for Minnesota's South Branch of the Root River Watershed. SWAT employs Hydrologic Response Units (HRUs) as elemental units, clustering grids by shared characteristics like soil type and elevation (see Figure \ref{fig:SWAT_model}). SW values average soil moisture both vertically within soil layers and horizontally across HRUs. SNO is calculated by averaging the remaining snow after accounting for melting factors. Streamflow (SF) measures water discharge at the watershed's outlet, routing water from upstream to downstream.

\begin{figure}[tb]
    \centering
    \includegraphics[width=\columnwidth]{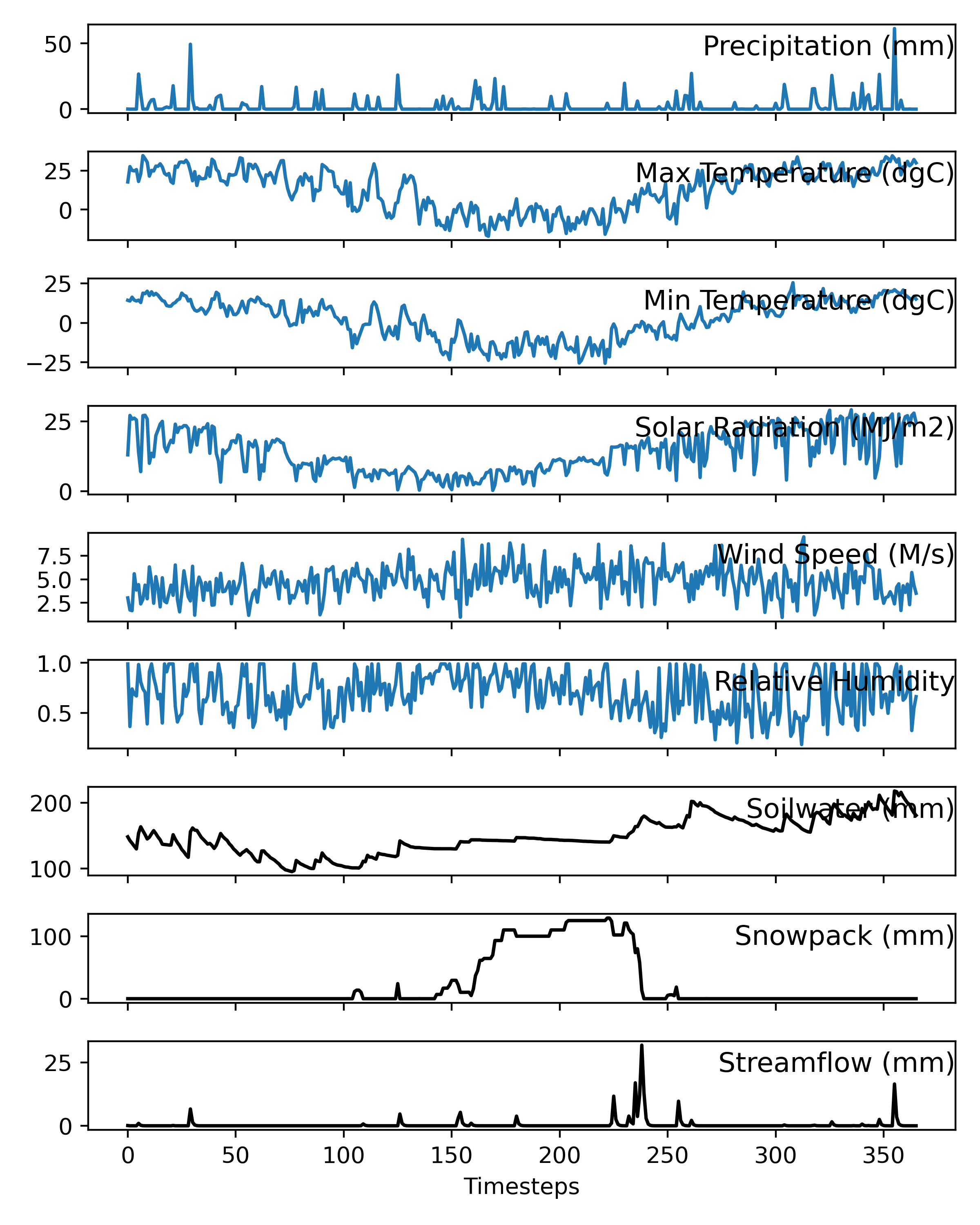}
    \caption{Six weather drivers and three responses in three SWAT datasets.}
    \label{fig:swat_data}
\end{figure}

The SW, SNO, and SF datasets, driven by six weather inputs, capture complex hydrological dependencies in a Minnesota watershed. As shown in Figure \ref{fig:swat_data}, SW demonstrates long-term, SNO exhibits seasonal, and SF has short-term dependencies, mostly driven by precipitation and temperature cycles. During winter, precipitation accumulates as SNO, keeping SW stable. Spring sees a surge in SF due to SNO melt, which also increases SW. Summer and autumn exhibit declining SF and SW levels as SNO depletes and evapotranspiration rises. These cyclical interactions present challenges for machine learning models in capturing the intricacies of the hydrological cycle, particularly in cold regions facing climate change impacts.

\subsection{Timestep-wise performance comparison.} \label{appendix:AvgStepRMSE}
\begin{figure}[tb]
\centering
\includegraphics[width=\columnwidth]{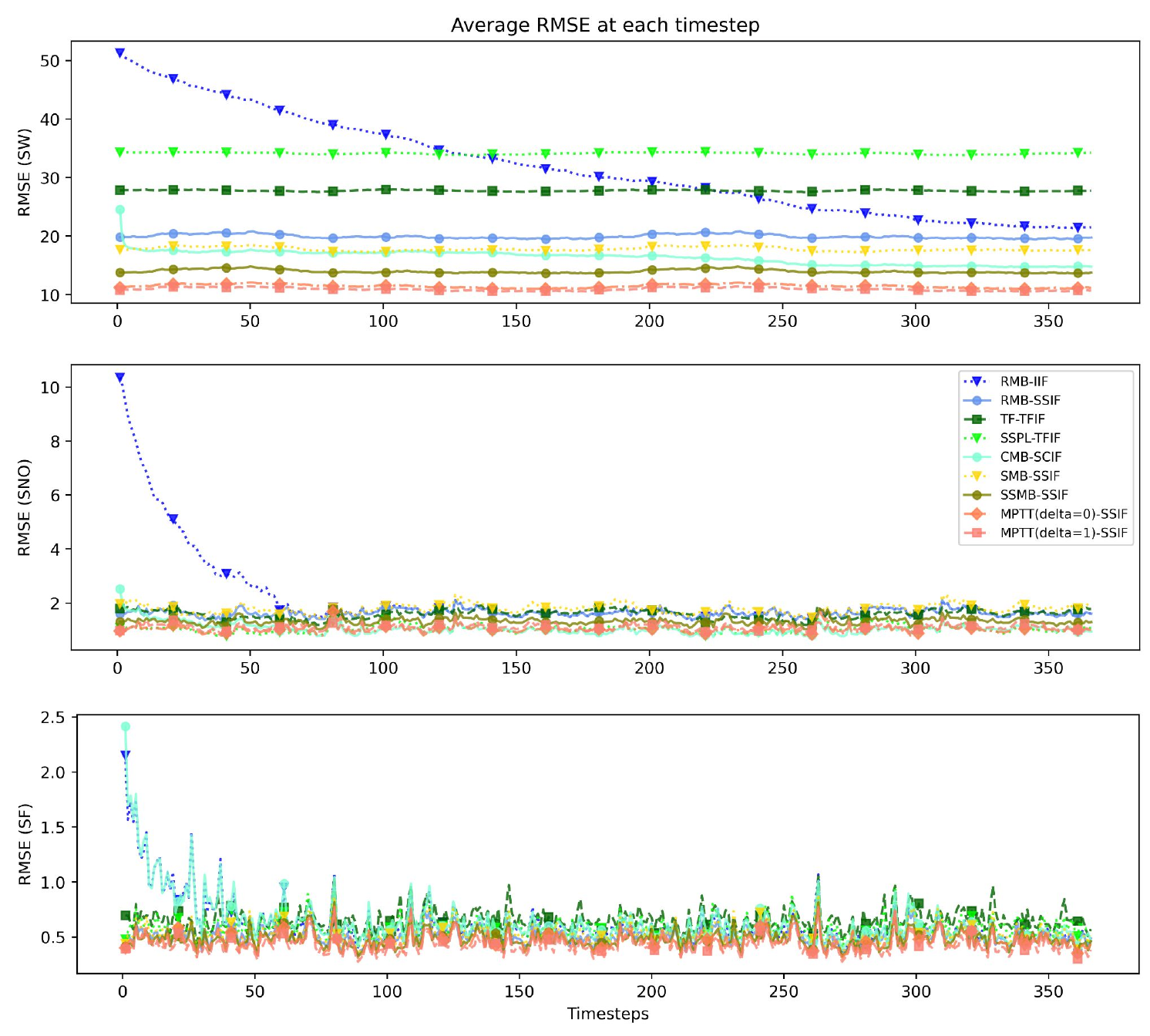}
\caption{This figure shows AvgStepRMSE traces across different learning strategies, highlighting the time steps needed for RNNs to form reliable hidden states essential for accurate inference.}
\label{fig:RMSEs_timestep}
\end{figure}
In this experiment, we partition each dataset into overlapping sequences of 366 steps, with a 183-timestep overlap. GRUs are trained across various strategies under uniform hyperparameters (see table \ref{tab:hyperparameters}). We then predict test sequences, calculate the RMSE for each timestep, and average RMSEs across all sequences at each timestep to generate AvgStepRMSE traces (see Figure \ref{fig:RMSEs_timestep}). This AvgStepRMSE metric offers insights into the RNN's capacity to form reliable hidden states for precise inference. Note that the first step of each sequence doesn't correspond to the start of the year due to overlaps and lunar years.

Figure \ref{fig:RMSEs_timestep} presents the following observations:
(a) Both RMB-IIF and CMB-SSIF require several timesteps to converge to lower AvgStepRMSEs since they initialize each RNN sequence with zero hidden states and must rebuild hidden states during inference.
(b) The SW and SNO plots demonstrate that CMB-SCIF converges to a lower AvgStepRMSE in far fewer steps than RMB-IIF. This rapid convergence is due to CMB-SCIF leveraging predicted initial responses to give trained RNNs a useful starting point to rebuild the hidden states. Conversely, CMB-SCIF and RMB-IIF show comparable performance on SF, as the short temporal dependency in SF implies that the responses hold limited information, offering little advantage to CMB-SCIF.
(c) Aside from RMB-IIF and CMB-SSIF, other strategies display flat AvgStepRMSE traces as they pass hidden states between sequences, eliminating the need for RNNs to rebuild hidden states during inference.
(d) In line with previous experiments, MPTT algorithms consistently yield the lowest AvgStepRMSEs and outperform baseline algorithms at each RNN timestep, reinforcing their superior performance.

\subsection{Season-wise performance comparison.}
\begin{figure}[tb]
\centering
\includegraphics[width=\columnwidth]{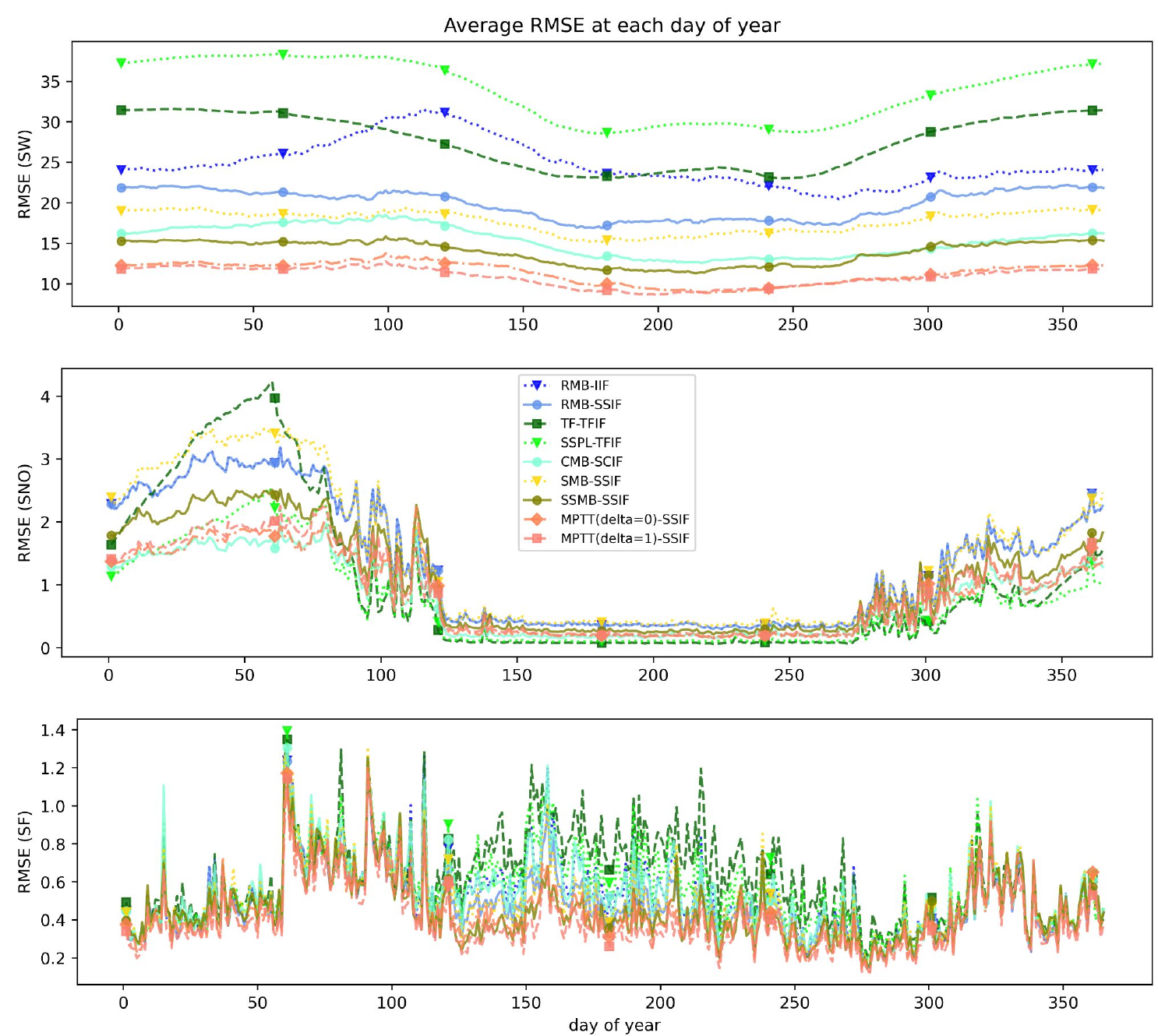} 
\caption{This figure displays AvgDailyRMSE trends for different learning strategies, highlighting seasonal variation in prediction accuracy for Soilwater (SW), Snowpack (SNO), and Streamflow (SF).}
\label{fig:RMSEs_DOY}
\end{figure}
In this experiment, we post-process predicted sequences back into time series form. We discard the first half of each overlapping sequence, retaining only the last 183 timesteps, and merge them to reconstruct the time series. This approach advantages strategies like RMB-IIF and CMB-SCIF, which perform better in sequence latter halves as indicated by AvgStepRMSE traces in Figure \ref{fig:RMSEs_timestep}. Subsequently, we calculate the AvgDailyRMSE by aligning RMSEs for each timestep according to the year's day and averaging these across multiple years, facilitating the analysis of seasonal RMSE fluctuations in the SW, SNO, and SF datasets.

Figure \ref{fig:RMSEs_DOY} highlights seasonal AvgDailyRMSE variations for SW, SNO, and SF. Elevated AvgDailyRMSEs occur in winter for SNO, driven by snowfall, and in spring for SW and SF due to snowmelt runoff. In contrast, summer and fall show lower AvgDailyRMSEs, reflecting reduced complexity as SNO depletes and both SW and SF become rain-dependent.

Figure \ref{fig:RMSEs_DOY} also elucidates specific performance trends: (a) As expected, TF-TFIF and SSPL-TFIF perform poorly on SW due to their error accumulation issue. These algorithms also underperform on SF, particularly during summer, similar to the RMB-IIF and CMB-SCIF strategies. (b) In winter, TF-TFIF and SSPL-TFIF underperform on the SNO dataset, showing a rising trend in AvgDailyRMSE due to error accumulation, as corroborated by Plot B of Figure \ref{fig:error_accumulate}. (c) Despite achieving acceptable overall RMSEs as per Table \ref{tab:SWAT} and Figure \ref{fig:winsize}, TF-TFIF and SSPL-TFIF's performance is skewed by lower AvgDailyRMSEs during summer and fall. In the Minnesota watershed studied, summer SNO levels are zero, making their decent SNO predictions in this period trivial. Consequently, TF-TFIF and SSPL-TFIF are not ideally suited for SNO modeling. (d) CMB-SCIF consistently excels over TF-TFIF and SSPL-TFIF, notably in mitigating error accumulation. Its stable AvgDailyRMSE during the snow season further highlights its effectiveness, contrasting with the increasing trend observed in TF-TFIF and SSPL-TFIF. (e) SSMB-SSIF consistently surpasses SMB-SSIF, particularly in SW predictions. This advantage stems from SSMB-SSIF's capability to preserve full sample and temporal dependencies during training, a capability SMB-SSIF lacks. (f) Figure \ref{fig:RMSEs_DOY} highlights MPTTs' consistent superiority in AvgDailyRMSE for SW and SF year-round while remaining competitive in SNO performance.

\section{Real world data experiments}\label{appendix:cvaravan-data}
\subsection{CARAVAN-GB dataset} \label{appendix:cvaravan-dataset}
CARAVAN is a publicly, comprehensive, real-world hydrology benchmark dataset for streamflow modeling~\cite{kratzert2022caravan}. It consolidates and standardizes seven existing large-sample hydrology datasets, including the US, Great Britain, Australia, Brazil, and Chile. Our study focuses on 191 Great Britain (GB) basins within CARAVAN, featuring meteorological drivers, streamflow data, and static characteristics like geophysical and climatological factors for each basin. Tables ~\ref{tab:forcing_data} and ~\ref{tab:basin_characteristics} outline the input drivers and basin characteristics used in our streamflow prediction experiment.

\begin{table}[ht]
    \resizebox{\linewidth}{!}{
        \begin{tabular}{|l|l|}
        \hline
        \textbf{Meteorological forcing data}                                               & \textbf{Unit} \\ \hline
        Daily Precipitation sum(total\_precipitation\_sum)                                 & $mm/day$        \\ \hline
        Daily Potential evaporation sum(potential\_evaporation\_sum)                       & $mm/day$        \\ \hline
        Mean Air temperature (temperature\_2m\_mean)                                       & $^\circ C$            \\ \hline
        Mean Dew point temperature (dewpoint\_temperature\_2m\_mean)                       & $^\circ C$            \\ \hline
        Mean Shortwave radiation (surface\_net\_solar\_radiation\_mean)                    & $Wm^{-2}$          \\ \hline
        Mean Net thermal radiation at the surface (surface\_net\_thermal\_radiation\_mean) & $Wm^{-2}$          \\ \hline
        Mean Surface pressure (surface\_pressure\_mean)                                    & $kPa$           \\ \hline
        Mean Eastward wind component (u\_component\_of\_wind\_10m\_mean)                   & $ms^{-1}$          \\ \hline
        Mean Northward wind component (v\_component\_of\_wind\_10m\_mean)                  & $ms^{-1}$         \\ \hline
        \end{tabular}
    }
    \caption{\small Summary of Meteorological Inputs for Streamflow Prediction.}
    \label{tab:forcing_data}
\end{table}

\begin{table}[ht]
\resizebox{\columnwidth}{!}{%
\begin{tabular}{|l|l|l|}
\hline
\textbf{Variable Name} & \textbf{Description} & \textbf{Unit}                                                                                                                                      \\ \hline
p\_mean                & Mean daily precipitation &  $mm/day$                                                                                                                                                \\ \hline
pet\_mean              & Mean daily potential evaporation &  $mm/day$                                                                                                                                          \\ \hline
aridity                & \begin{tabular}[c]{@{}l@{}}Aridity index, ratio of \\mean PET and mean precipitation\end{tabular} & --                                                                                                                   \\ \hline
frac\_snow             & Fraction of precipitation falling as snow     & --                                                                                                                             \\ \hline
moisture\_index        & Mean annual moisture index in range {[}-1, 1{]}  &--                                                                                                                        \\ \hline
seasonality            & Moisture index seasonality in range {[}0, 2{]} &--                                                                                                                            \\ \hline
high\_prec\_freq       & \begin{tabular}[c]{@{}l@{}}Frequency of high precipitation days, where\\ precipitation $\geq$ 5 times mean daily precipitation\end{tabular}  &--                                   \\ \hline
high\_prec\_dur        & \begin{tabular}[c]{@{}l@{}}Average duration of high precipitation events\\ (number of consecutive days where \\precipitation $\geq$ 5 times mean daily precipitation\end{tabular} & days \\ \hline
low\_prec\_freq        & Frequency of low precipitation days     & --                                                                                                                                   \\ \hline
low\_prec\_dur         & The average duration of low precipitation events     & days                                                                                                                      \\ \hline
kar\_pc\_sse           & Karst area extent    & \% cover                                                                                                                                                      \\ \hline
cly\_pc\_sav           & Clay fraction in soil    & \%                                                                                                                                                  \\ \hline
slt\_pc\_sav           & Silt fraction in soil   & \%                                                                                                                                                   \\ \hline
snd\_pc\_sav           & Sand fraction in soil  & \%                                                                                                                                                    \\ \hline
soc\_th\_sav           & Organic carbon content in soil  & tonnes/hectare                                                                                                                                           \\ \hline
swc\_pc\_syr           & Annual mean soil water content       & \%                                                                                                                                      \\ \hline
ele\_mt\_sav           & Elevation     & $m$ above sea level                                                                                                                                                             \\ \hline
slp\_dg\_sav           & Terrain slope  & °(x10)                                                                                                                                                            \\ \hline
basin\_area            & Basin Area   &      $km^2$                                                                                                                                                        \\ \hline
for\_pc\_sse           & Forest cover extent    & \% cover                                                                                                                                                    \\ \hline
\end{tabular}%
}
\caption{Summary of Basin Characteristics Inputs for Streamflow Prediction.}
\label{tab:basin_characteristics}
\end{table}

\subsection{A closer look at algorithm performance}
\begin{figure}[tb]
\centering
\includegraphics[width=\columnwidth]{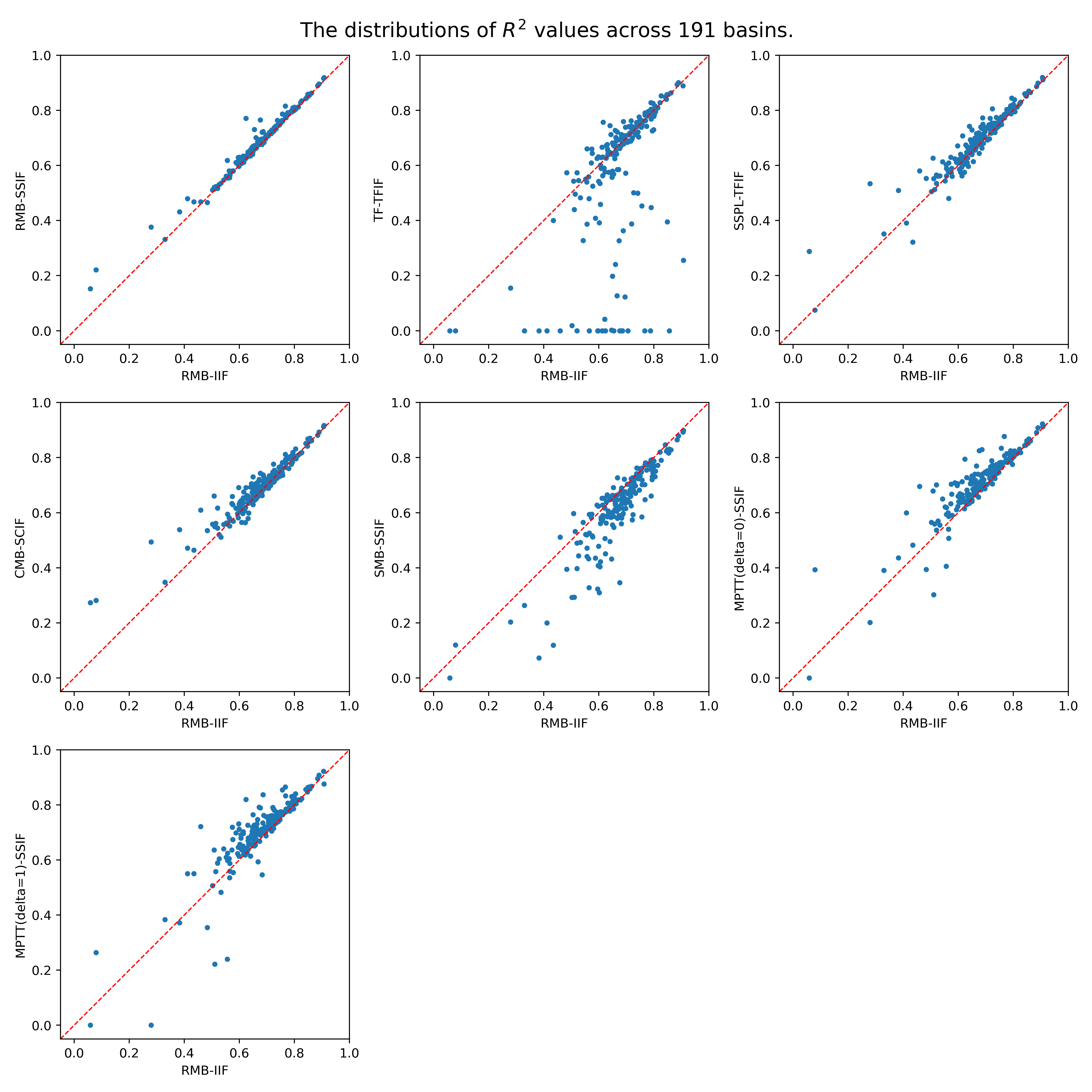} 
\caption{Scatter plots of $R^2$ distributions between RMB-IIF and other learning strategies for 191 basins in the GB, with negative $R^2$ values set to zero.}
\label{fig:caravan_scatter_R2}
\end{figure}
Figure \ref{fig:caravan_scatter_R2} depicts the $R^2$ score distribution between RMB-IIF and other learning strategies using scatter plots; negative $R^2$ scores are set to zero for clarity. Key observations include:
(a) Using the same LSTMs as RMB-IIF, RMB-SSIF outperforms it in 173 of 191 basins. The simple switch from IIF to SSIF inference yields consistent performance gains, highlighting its utility for improving existing models in other applications.
(b) TF-TFIF outperforms RMB-IIF in just 83 of 191 basins and produces negative $R^2$ values in 22 basins.
(c) SSPL-TFIF surpasses TF-TFIF and shows $R^2$ values akin to CMB-SCIF. However, CMB-SCIF remains the preferred choice due to its lower error accumulation, as elaborated in section \ref{sec:SWAT_exps}.
(d) MPTT strategies notably enhance $R^2$ values across multiple basins, with MPTT(delta=1)-SSIF outperforming RMB-IIF in 155 of 191 basins.

\subsection{Predicted timeseries}
\vspace{-0.5\baselineskip}
\begin{figure}[htbp]
\centering
\includegraphics[width=\columnwidth]{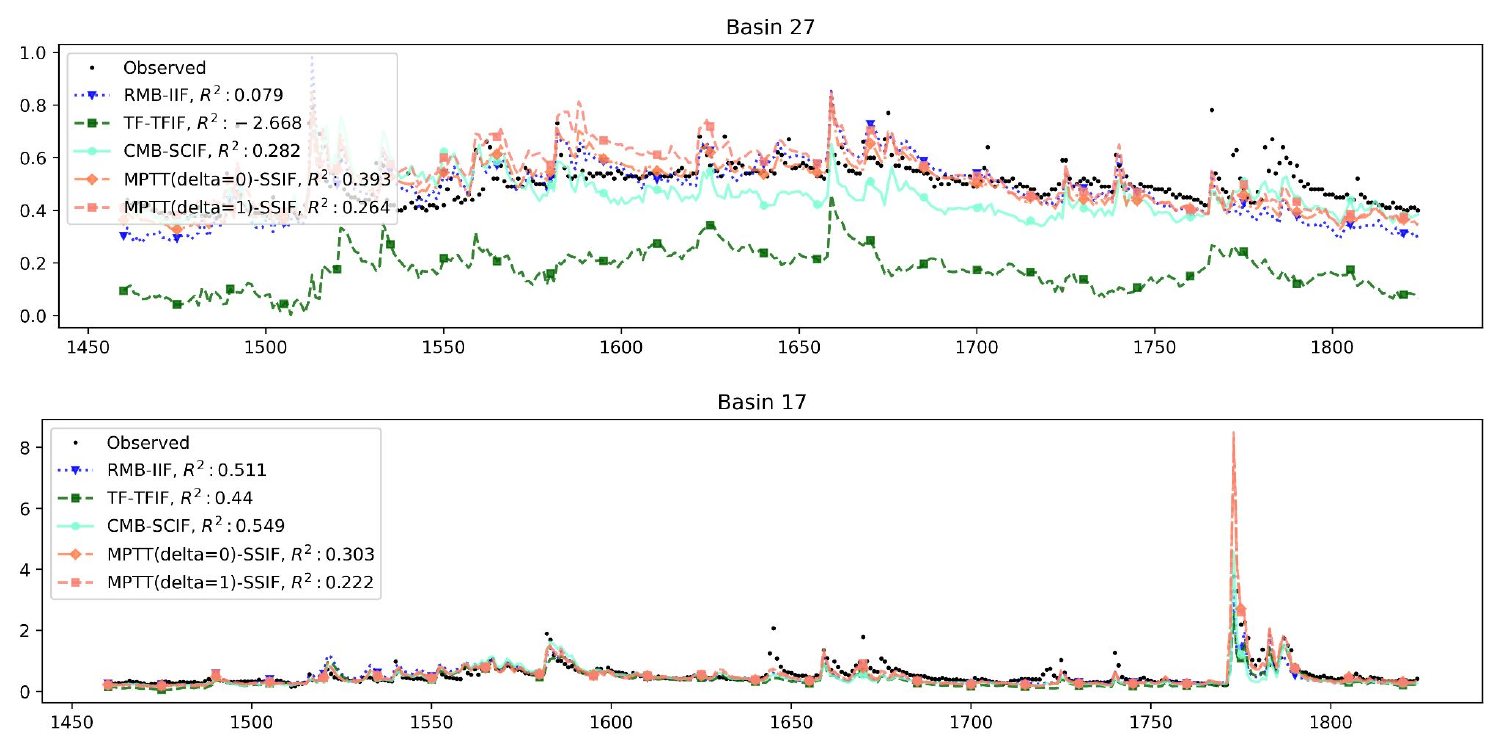} 
\caption{Visual Comparisons of Best and Worst MPTT-SSIF Performances Versus RMB-IIF in Basins 27 and 17.}
\label{fig:caravan_sequences}
\end{figure}
In the experiment, MPTT-SSIF outperforms RMB-IIF in 155 of 191 basins in terms of $R^2$ values. We calculate and rank the $R^2$ differences between MPTT-SSIF and RMB-IIF, selecting the best (basin 27) and worst (basin 17) performing basins of MPTT-SSIF relative to RMB-IIF for further analysis. These basins' time series are visualized in Figure \ref{fig:caravan_sequences}. We additionally visualize TF-TFIF and CMB-SCIF to scrutinize TF-TFIF's frequent negative $R^2$ scores.

MPTT-SSIF exhibits its most marked improvement over RMB-IIF in basin 27, characterized by a stable flow and extended temporal dependencies, as indicated in Figure \ref{fig:caravan_scatter_R2}. As MPTT-SSIF and CMB-SCIF are adept at maintaining both sample and temporal relationships, they naturally outperform RMB-IIF. Specifically, MPTT-SSIF's predictions align closely with actual data during both stable and peak flow conditions, leading to elevated $R^2$ scores. Although the trend of TF-TFIF's predicted time series generally follows the actual data, a significant discrepancy exists due to its issue with error accumulation.

Despite being the least effective against RMB-IIF on basin 17, MPTT-SSIF still manages an $R^2$ of 0.303, nearly on par with the 0.5 $R^2$ achieved by other strategies. Figure \ref{fig:caravan_scatter_R2} indicates close alignment between predicted and actual time series for all four methods, confirming their comparable performance on this basin. As discussed in Section \ref{sec:SWAT_exps}, learning strategies tend to perform similarly on systems with short-term temporal dependencies, a characteristic exemplified by basin 17. The sharp peaks in the data disrupt the temporal patterns, causing the learning strategies to perform similarly.

\section{Pseudocodes} \label{appendix:pseudocodes}
\begin{algorithm}[ht] \label{alg:mptt}
\caption{Message Propagation Through Time (MPTT) Training }
\label{alg:MPTT}
\begin{flushleft}
\textbf{Input}: Training sequences $\{S_1, S_2,\dots, S_M\}$, sequence IDs $\{\text{id}_1, \text{id}_2, \dots, \text{id}_M\}$, and message keeper $\delta$.
\end{flushleft}
\begin{algorithmic}[1] 
\STATE Create a \emph{key-map},  following equation \ref{eq:key_map}.
\STATE Initialize $\emph{state-map}(k)=(\vec{0}, \vec{0}, 0)$ for each key $k$, following equation \ref{eq:state_map}.
\WHILE{maximum epochs not reached}
\STATE Randomly partition the sequences into multiple mini-batches without replacement.
\FOR{each mini-batch}
    \FOR{each sequence $S$ in the mini-batch}
        \STATE Retrieve $(\mu_0, \bar{h}^0, c)$ from \emph{state-map} corresponding to the key of S, and generate $\tilde{h}_0$ using equation \ref{eq:MPTT_fetch_policy} to initialize the hidden state of S, following the \emph{Read policy}.
    \ENDFOR
    \STATE Obtain the predictions and hidden states at each time step for the mini-batch of sequences using equation \ref{eq:MPTT_hs}.
    \STATE Compute the loss and update the model.
    \STATE Detach hidden states.
    \FOR{each sequence $S$ in the mini-batch}
        \STATE Restrive a list of keys from the \emph{key-map} corresponding to the ID of $S$ 
        \FOR{each key in the list of keys}
            \STATE Extract the corresponding hidden state $h$ from the detached hidden states of $S$ using the key, following the \emph{write policy}.
            \STATE Retrieve $(\mu_0, \bar{h}_0, c)$ from $\emph{state-map}\text{(key)}$
            \STATE Update the values of $\bar{h}_0$ and c as $\bar{h}_0 = \frac{c\bar{h}_0 + h}{c+1}$ and $c = c+1$, respectively (equation \ref{eq:MPTT_update_policy}). 
            \STATE Update $\emph{state-map}\text{(key)}$ with the new value $(\mu_0, \bar{h}_0, c)$.
        \ENDFOR
    \ENDFOR
\ENDFOR
\FOR{each key in \emph{state-map}}
    \STATE Retrieve the value $(\mu_0, \bar{h}_0, c)$ from $\emph{state-map}(\text{key})$.
    \IF{$c\geq 1$}
    \STATE Compute the new $\mu_0 = \frac{\delta \mu_0  + c\bar{h}_0}{\delta + c}$, $\bar{h}_0 = \vec{0}$ and $c = 0$ (equation \ref{eq:MPTT_pass_policy}).
    \STATE Update $\emph{state-map}(\text{key}) = (\mu_0, \bar{h}_0, c)$, following the \emph{Propagate policy}.
    \ENDIF
\ENDFOR
\ENDWHILE
\end{algorithmic}
\end{algorithm}

\end{document}